\documentclass[11pt,letterpaper]{article}

\usepackage[letterpaper, margin=1in]{geometry}

\usepackage{setspace}
\usepackage{amssymb,amsmath}
\usepackage{amsthm}
\usepackage{thmtools}
\usepackage{thm-restate}

\usepackage{algorithm,algorithmicx}
\usepackage{algpseudocode}

\usepackage{tikz}
\usepackage{graphicx}
\usepackage{caption}
\usepackage{subcaption}
\usepackage{float}

\usepackage{booktabs}
\usepackage{multirow}
\usepackage{multicol}

\usepackage{bbm}
\usepackage{latexsym}
\usepackage{mathrsfs}
\usepackage{xcolor}
\usepackage{amssymb}

\usepackage{enumerate}
\usepackage{breqn}
\usepackage{xspace}
\usepackage{url}
\usepackage{hyperref}
\usepackage{cleveref}

\usepackage[numbers]{natbib}
\bibpunct[, ]{(}{)}{,}{n}{}{,}
\newtheorem{theorem}{Theorem}
\newtheorem{lemma}{Lemma}
\newtheorem{definition}{Definition}
 
 \newtheorem{remark}[theorem]{Remark}

\newcommand{\bD}{\mathbf{D}}
\newcommand{\bE}{\mathbf{E}}

\newcommand{\pr}{\Pr}
\newcommand{\E}{\mathbb{E}}

\newcommand{\R}{\mathbb{R}}

\newcommand{\I}{\mathcal{I}}

\newcommand{\cC}{\mathcal{C}}

\newcommand{\cP}{\mathcal{P}}
\newcommand{\calE}{\mathcal{E}}

\newcommand{\OPT}{\mathsf{OPT}}

\newcommand{\ALG}{\mathtt{ALG}}
\newcommand{\calG}{\mathcal{G}}
\newcommand{\TV}{\operatorname{\mathsf{TV}-\mathsf{Dist}}}
\newcommand{\ESD}{\mathtt{EmpStocDom}}

\newcommand{\hF}{G}

\newcommand{\bld}[1]{\mathbf{{#1}}}

\newcommand{\varX}{\mathsf{X}}
\newcommand{\varY}{\mathsf{Y}}

\newcommand{\bx}{\mathbf{x}}

\newcommand{\calD}{\mathcal{D}}
\newcommand{\calI}{\mathcal{I}}

\newcommand{\1}{\mathbf{I}}

\def\sd{\preceq_{\mathsf{SD}}}

\def\calI{\mathcal{I}}

\newcommand{\TO}{\tilde{O}}

\def\H{\mathrm{H}}
\def\supp{\mathrm{supp}}

\title{{Near-Optimal Regret for Semi-Bandit Learning of \\ Monotone Stochastic Problems}}
\begin{document}

\author{Zhengjia Zhuo\thanks{Department of Industrial and Operations Engineering, University of Michigan, Ann Arbor, USA. Research supported in part by NSF grant  CCF-2418495.}  \and Viswanath Nagarajan$^*$}

\maketitle

\begin{abstract}
     We consider the task of learning an (approximately) optimal policy for stochastic optimization problems where the true distributions are unknown. In each period, the algorithm plays a policy for the stochastic problem and observes semi-bandit feedback, i.e., it {\em only} receives  new samples from the random variables that were selected by the policy.
     We obtain an   algorithm that works for a  large class of ``monotone'' stochastic problems, including prophet inequality, Pandora's box, series testing, stochastic knapsack and stochastic matching. We obtain  $\TO(\sqrt{nkT})$ regret for ``weakly'' monotone problems and    $\TO(\sqrt{nT})$ regret
    for ``strongly'' monotone problems, where  $T$ is the number of periods, $n$ is the problem dimension (number of random variables) and $k$ is the maximum support size of any random variable.
          This  improves over previous regret bounds by a $\sqrt{n}$ factor. In fact, our dependence on $n$ and $T$ is the best-possible due the $\Omega(\sqrt{nT})$ lower bound for multi-armed-bandits.
     Our main  technique is the use of  Hellinger distance
     in bounding the ``error'' between the true and learned distributions, and the use of a constant-factor chain rule for the Hellinger distance of dependent random variables. \end{abstract}

\section{Introduction}
Stochastic optimization involves problems where certain input parameters are random variables (r.v.s)  specified via their probability distributions. These arise in various settings such as optimal stopping theory~\citep{weitzman1979,samuel1984comparison,kleinberg2012matroid}, mechanism design~\citep{Myerson81,ChawlaHMS10} and stochastic combinatorial optimization \citep{deshpande2016approximation,DGV08,ChenIKMR09,BGLMNR12,GN13,deshpande2016approximation}. While several elegant results (exact as well as approximation algorithms) are known for  these problems, they rely crucially on knowing the underlying distributions.

Recent  work has considered relaxing the known-distributions assumption in several stochastic optimization problems. One line of work \citep{cole2014sample, roughgarden2016ironing,GuoHZ19,GuoHT+21} provides sample-complexity bounds on the number of   samples (from all r.v.s) needed to obtain an $\epsilon$-additive approximation to the optimal value.  Another line of work \citep{lattimore2020bandit,GatmiryKS+22,agarwal2023semi,liu2025improved} provides regret bounds for online (semi)bandit learning
in a $T$-period setting,
where the algorithm faces the same (unknown) distribution in each period. Here, in each period the learning algorithm only observes samples from the r.v.s that are actually ``probed'' by the current solution.

Our main contribution is a near-optimal regret bound  for semi-bandit learning for a large class of ``monotone'' stochastic problems. In particular, we obtain $\tilde{O}(\sqrt{nkT})$ regret for any weakly-monotone problem, where $n$ is problem dimension (number of r.v.s), $k$ is the maximum support size of any r.v. and $T$ is the number of periods. For the more restricted class of strongly-monotone problems, we obtain an improved    $\tilde{O}(\sqrt{nT})$ regret bound, which is independent of the support. These results improve prior regret bounds by a $\sqrt{n}$ factor, and are nearly optimal due to an $\Omega(\sqrt{nT})$ lower bound for the basic multi-armed-bandit problem \citep{lattimore2020bandit}.

\subsection{Problem Definition}
We consider the same problem setting as \cite{agarwal2023semi}, which captures fundamental stochastic problems such as  series testing, Pandora's box, stochastic knapsack and stochastic matching.
\paragraph{Traditional Stochastic Optimization} A stochastic problem  $\cP$ consists of  $n$ items with an independent real-valued random variable (r.v.) $\varX_i$  associated with each item $i\in \calI$.
There is a \emph{known} probability distribution $\calD_i$ for $\varX_i$. We assume that each $\calD_i$ has finite support of size at most $k$. We use $\bD=\calD_1\times \cdots \times \calD_n$ to denote the joint distribution over all items.

A solution or  {\em policy} $\sigma$  for $\cP$  is given by a decision tree $T_\sigma$ along with a cost at each leaf-node of $T_\sigma$. Each internal node of $T_\sigma$ is labeled by an item to probe next, and the branches out of such a node correspond to the random realization of the probed item. Leaf nodes in $T_\sigma$ correspond to  termination decisions of the policy, and each such node $l$  is also associated with a non-negative cost $f_{\sigma,l}$. We assume a uniform upper bound $f_{max}$ on the cost at any leaf of any policy.   Each item may be probed at most once in a root-leaf path. Nodes in the decision tree  can also be viewed as    \emph{states}   of the policy, which correspond to the sequence of previously-probed items  with their realizations. The root node of $T_\sigma$  is the starting state of the policy, at which point no item has been probed.  The {\em policy execution} under any realization corresponds to   a unique root-leaf path in $T_\sigma$, and the cost incurred is the $f_{\sigma,l}$ value at the  leaf $l$. We also allow for constraints in problem  $\cP$, which restrict the set of allowed decision trees and leaf costs:  $\cC$ denotes the set of feasible policies.  We use $f(\sigma|\bD)$ to denote the expected cost of policy $\sigma$ under distribution $\bD$.  We also work with {\em randomized} policies, where each internal node  in the decision tree corresponds to a  probability distribution  over  items.
Our goal is to find a policy $\sigma$ that minimizes (or maximizes) the expected cost under $\bD$. $$    \text{minimize} \quad f(\sigma|\bD) \quad  \text{subject to} \quad \sigma \in \cC.$$
We use $\sigma^*$ to denote the optimal policy and $\OPT(\bD)=f(\sigma^*|\bD) $ the optimal value.

It is worth noting that any such stochastic problem can be solved optimally as a Markov Decision Process (MDP). However, this is not a computationally efficient approach because the number of states in the MDP is exponential in the problem dimension $n$. An alternative approach, which has been very successful, is via (polynomial-time) approximation algorithms  for  stochastic problems.
An $\alpha$-approximation algorithm for a
stochastic problem $\cP$ (with minimization objective) takes as input the probability distributions $\{\calD_i\}_{i \in \calI}$ along with any other objective/constraint parameters, and returns a policy that has expected cost at most $\alpha\cdot \OPT(\bD)$.

\paragraph{Online Semi-Bandit Learning.}
In this setting, the distributions $\calD_1, \ldots, \calD_n$ are {\em unknown}; we assume that other parameters such as the cost function and constraints are known. We are interested in a learning algorithm that converges (in objective value) to the optimal policy by     repeatedly solving the  stochastic problem $\cP$
over $T$ time periods.
The underlying distributions $\{\calD_i\}_{i \in \calI}$ remain the same across  periods.
Formally, in each period  $t \leq T$, the algorithm  uses all prior observations to compute policy $\sigma^t$. We measure the learning algorithm in terms of \emph{expected total regret}.  For minimization problems, this is:  \begin{align*}
    R(T) =  \sum_{t=1}^T \E_{}\left[  f(\sigma^t |\bD)\right] -  T\cdot \OPT(\bD).
\end{align*}
where the expectation is taken over all randomness. Importantly, we consider  \emph{semi-bandit}  feedback, where   in each period  $t$, the algorithm only observes the set of items probed by policy $\sigma^t$ before it terminates. Semi-bandit feedback is more realistic than ``full feedback'' where the learning algorithm observes one sample from {\em every} item in each round.  As observed  in \cite{agarwal2023semi}, learning under full-feedback is closely related to sample-complexity bounds, whereas semi-bandit learning requires significantly more work.

In order to allow for computationally efficient learning algorithms, we use the notion of $\alpha$-regret, defined as follows.
\begin{equation*}
      \sum_{t=1}^T \E_{}\left[  f(\sigma^t |\bD)\right]- \alpha \cdot T\cdot  \OPT(\bD) .
\end{equation*}

\def\sp{ \Gamma}

\def\sd{\preceq_{\mathsf{SD}}}

\subsection{Results and Techniques}
Before stating our results formally, we need some definitions.

\begin{definition}[Stochastic Dominance]\label{def:stoch-dom}
A probability distribution $\calE$ (over $\R$)  stochastically dominates another distribution $\calD$ if, for all $a\in \R$,  we have: $\Pr_{\varX \sim \calE}  (\varX  \geq a) \geq \Pr_{\varY \sim \calD}(\varY \geq a)$.
\end{definition}

We use $\calD \sd \calE$ to denote that distribution  $\calE$  stochastically dominates $\calD$.
For product distributions  $\bld{D} = \{\calD_i\}_{i=1}^n$ and $\bld{E}= \{\calE_i\}_{i=1}^n$, we say that $\bld{E}$  stochastically dominates $\bld{D}$ if $\calE_i$ stochastically dominates $\calD_i$ for all $i\in [n]$; we also denote this by $\bld{D}\sd  \bld{E}$.

For  any stochastic problem instance $\sp$ and distribution $\bld{D}$, let $\OPT_{\sp}(\bld{D})$  denote the optimal cost of instance $\sp$ with r.v.s having distribution $\bld{D}$. We work with two notions of stochastic monotonicity, that have also been used in prior work \cite{GuoHZ19,GuoHT+21,agarwal2023semi}.
\begin{definition}[Weak Monotonicity]\label{def:weak monotonicity} A stochastic problem  $\cP$
is {\em weakly up-monotone} if for any instance $\sp$ and probability distributions $\bld{D}$ and $\bld{E}$  where $\bld{D} \sd \bld{E}$, we have $\OPT_\sp(\bld{E})\le  \OPT_\sp(\bld{D})$. Similarly, the problem is said to be {\em weakly down-monotone} if for any instance $\sp$ and probability distributions $\bld{D}$ and $\bld{E}$  where $\bld{D} \sd \bld{E}$, we have $\OPT_\sp(\bld{E})\ge  \OPT_\sp(\bld{D})$.
\end{definition}

\begin{definition}[Strong Monotonicity]\label{def:strong monotonicity} A stochastic problem $\cP$
is  strongly up-monotone if for any instance $\sp$ and probability distributions $\bld{D}$ and $\bld{E}$  where $\bld{D} \sd \bld{E}$, we have $f(\sigma^*_{\bD}| \bE)\le  \OPT_\sp(\bld{D}) = f(\sigma^*_{\bD}| \bD)$, where $\sigma^*_{\bD}$ is the optimal policy under $\bD$. Strongly down-monotone is defined analogously.
\end{definition}
Clearly, strong monotonicity implies weak monotonicity. In this paper, we focus on weak/strong up-monotone problems with a minimization objective; all our results extend in a straightforward way (by symmetric arguments) to down-monotone problems and/or maximization objective.

Our first main result is:
\begin{theorem}\label{thm:main strong-discrete}
Suppose that a stochastic problem $\cP$ with discrete distributions has an  $\alpha$-approximation algorithm, and  is weakly monotone. Then, there is a  polynomial time semi-bandit  learning algorithm  for $\cP$
with $\alpha$-regret $O\left( \left(\sqrt{n  kT} \log^{\frac32} (T)+n\log(T)\right ) f_{\max} \right)$. Here, $n$ is the number of items, $k$ is the maximum support size of any distribution, $f_{\max}$ is the maximal value of the objective function,
and $T$ is the number of time periods.
\end{theorem}
This result improves upon the $\tilde{O}\left(n\sqrt{kT} \, f_{max}\right)$ regret bound for the same problem class  \cite{agarwal2023semi}. As an immediate consequence, we obtain improved $\alpha$-regret bounds for many  stochastic problems such as stochastic knapsack,  matching, orienteering, probing, submodular maximization  etc; in each case we  improve regret by a $\sqrt{n}$ factor.  Moreover, we obtain an optimal dependence on $n$ and $T$ (up to poly-log factors). This is because the classic multi-armed-bandit problem (which is a very special case) has a tight $\tilde{\Theta}(\sqrt{nT})$ regret bound.

Our algorithm is similar to that of \cite{agarwal2023semi}. In each period, (1) we construct a shifted empirical distribution $\calE_i$ for each item $i$ that dominates the true distribution $\calD_i$ and (2) we run the approximation algorithm using the known distributions $\{\calE_i\}_{i=1}^n$. We also follow the same high-level analysis. The key novelty lies in establishing a stronger ``stability'' bound in terms of the Hellinger-distance between $\calD_i$ and  $\calE_i$. Specifically, given any policy $\sigma$ and two product distributions $\bD$ and $\bE$, we prove in Lemma~\ref{lem:stb:weak} that:
$$    f(\sigma | \bD) - f(\sigma | \bE) \,\,\lesssim   \,\, f_{\max} \cdot \sqrt{ \sum_{i \in [n]} {Q}_{i}(\sigma) \cdot\H^2(\calD_i,  \calE_i)  } ,
    $$
    where $Q_i(\sigma)$ is the probability that policy $\sigma$ probes item $i$ and $\H^2$ is the squared-Hellinger distance. This proof utilizes an adaptive chain rule for Hellinger distance due to \cite{Jayram2009} (see Theorem~\ref{lem:Jayram}), which turns out be simple in hindsight. Previously, \cite{agarwal2023semi} used the total-variation distance in this stability bound, which  leads to an $\TO(n\sqrt{T})$ regret.

    The above stability bound is used in each period,  and to bound the total regret  we rely crucially on the fact that an item $i$ contributes to this bound only when it is  probed (this is due to  the $Q_i(\sigma)$  term). Using Jensen's inequality, we can  then bound the total  regret
    as a function of the total $\H^2$-error over all periods and items. For any single item, the $\H^2$-error over all periods can be bounded by roughly $k\sum_{t=1}^T \frac1t\approx k \log T$, which leads to the final $\TO(\sqrt{nkT})$ bound.

When the stochastic problem is strongly monotone and there is an exact algorithm, we obtain an even better regret bound, which does not depend on the support.
\begin{theorem}\label{thm:main strong-general}
Suppose that a stochastic problem $\cP$ with general distributions has an exact algorithm
and   is strongly monotone. Then, there is a semi-bandit  learning algorithm  for $\cP$ with $1$-regret
$O\left(  \left(\sqrt{nT} \log^{\frac32} T+n\log T\right ) f_{\max} \right)$. Here, $n$ is the number of items, $f_{\max}$ is the maximal value of the objective function and $T$ is the number of time periods.
\end{theorem}

\begin{remark}\label{remark:alpha-monotone}

It is also
worth noting that the exact oracle assumption in Theorem~\ref{thm:main strong-general} can be relaxed under an additional
policy-wise monotonicity assumption. In particular, suppose the stochastic problem is weakly monotone  and  the policy returned by an $\alpha$-approximation oracle is itself
monotone, i.e., if $\sigma$ is the approximate policy returned by the oracle for distribution $\mathbf{E}$ then
\begin{equation}\label{eq:policy-wise-monotone}
f(\sigma\mid \hat{\mathbf{D}})
\leq
f(\sigma\mid\mathbf{E})
\qquad
\text{whenever }
\mathbf{E}\preceq_{\mathsf{SD}} \hat{\mathbf{D}}.
\end{equation}
With this extra assumption, we can obtain $\alpha$-regret $O\left(  \left(\sqrt{nT} \log^{\frac32} T+n\log T\right ) f_{\max} \right)$.
\end{remark}

This bound is optimal up to logarithmic terms (as noted above). As  applications, we obtain semi-bandit learning algorithms with $\TO(\sqrt{nT})$ regret for   prophet inequality, Pandora's box and sequential posted pricing. For the specific cases of   prophet inequality and Pandora's box,  \cite{liu2025improved} also obtained these optimal regret bounds. Our approach is more general as it directly applies to any strongly monotone problem.

The improved bound for strongly monotone problems relies on applying the stability lemma to the true distribution and a ``shifted true'' distribution (instead of the shifted empirical distribution as before). Intuitively, this shifted distribution is constructed by translating the cumulative distribution function of the true distribution. This type of shifting provides significantly more control in the analysis, as both distributions share the same support. We also need to handle some additional  technical issues in proving the  stability lemma for arbitrary distributions (instead of discrete).
To address this, we use Radon-Nikodym derivatives from measure theory (see Appendix~\ref{app:standard results} for measure-theoretic preliminaries).

\subsection{Related Work}
While stochastic optimization problems (known distributions) have been studied extensively for several decades, the learning setting has only been considered recently; see e.g., \cite{GatmiryKS+22,agarwal2023semi,SW2024,liu2025improved}
Any stochastic problem over  discrete distributions can be viewed as an MDP. While there are semi-bandit learning algorithms for MDPs, their regret bounds are  $\tilde \Theta(\sqrt{ST})$, where $S$ is the size of the state space \citep{OsbandR16b,AzarOM17}. When applied to stochastic problems with $n$ r.v.s, the state space is usually exponential in $n$. So, MDP learning results are insufficient to obtain regret bounds that are polynomial in $n$.  Recent work on learning for stochastic optimization, including this paper, can be seen as ways to utilize problem structure to bypass the  general MDP lower-bound.

The work closest to ours is \cite{agarwal2023semi}, where they obtained semi-bandit learning algorithms  for weakly monotone stochastic problems, and provided several applications. The original paper \cite{AgarwalGN24} obtained $\TO(nk\sqrt{T})$ regret, which was improved to  $\TO(n\sqrt{kT})$ regret in \cite{agarwal2023semi}.
We improve their bound by  a $\sqrt{n}$ factor.
While \cite{liu2025improved} achieved an optimal regret bound of $\TO(\sqrt{nT})$ for  both  prophet inequality and Pandora's box, their approach is highly problem-specific. They analyze the structural properties of the reward function to establish its $1$-Lipschitz continuity, relying on a change of variables for Lipschitz functions and the Kolmogorov-Smirnov (K-S) distance between two distributions. Furthermore, their framework is heavily dependent on the specific structure of the optimal policy---namely, that the optimal threshold is non-increasing across items. In contrast, our approach for Theorem~\ref{thm:main strong-general} is to  bound  the Hellinger distance and leverage general properties of strongly monotone problems, thereby completely bypassing the need for such problem-specific  analysis.

As mentioned earlier, there is also a line of work that obtains
sample complexity
bounds for  stochastic optimization problems.
The goal here is to learn an $\epsilon$-additive approximation to the optimal value using $poly(n,\frac1\epsilon)$ samples; each sample corresponds to one draw from $\calD_1\times \cdots \times \calD_n$. The work of ~\cite{GuoHT+21} gives
optimal sample complexity bounds for strongly monotone stochastic optimization problems. Their techniques also rely on proving bounds on the Hellinger distance:  we can essentially use their sampling result directly to construct our shifted empirical distribution.   We note that the connection between sample-complexity and learning under ``full feedback'' is well-known: see \cite{agarwal2023semi} for a formal argument that proves $\TO(\sqrt{nT})$ regret for monotone stochastic problem.  However, the only  generic way to use sample-complexity bounds for semi-bandit learning is via an explore-then-commit approach that incurs $\tilde\Omega(T^{2/3})$ regret, which is sub-optimal in $T$. Our approach can be seen as   using sample-complexity techniques (in a non-black-box manner) to prove semi-bandit regret  bounds: the key ingredient for this is  the adaptive chain rule for $\H^2$ distance \cite{Jayram2009}.

\section{Preliminaries}
We present some preliminaries that are required in our algorithm and proofs. The following results for Hellinger distance can be found at \cite{Gibbs2002Choosing}.
\begin{definition}\label{def:Hellinger}
    Given two discrete distributions $P = (p_1 \dots p_k)$ and $Q=(q_1\dots q_k)$, the Hellinger square distance is defined by
    $$
    \H^2(P,Q) = \frac{1}{2}\sum_{i\in [k]} \left(\sqrt{p_i} -\sqrt{q_i}\right)^2 .
    $$
    And for general distributions $P$ and $Q$ over support $S$ that are both absolute continuous with a common measure $\lambda$, we have
        $$
    \H^2(P,Q) = \frac{1}{2}\int_{S} \left(\sqrt{dP/d\lambda} -\sqrt{dQ/d\lambda}\right)^2 .
    $$
    Moreover, Hellinger distance is a valid metric: it satisfies triangle inequality.
\end{definition}

\begin{lemma}\label{lem:function-val-tv}
Given  product
distributions $\bld{D}$ and $\bld{E}$ over $n$ random variables $\varX_1, \ldots, \varX_n$,   and a $[0,U]$ bounded function $h$ on the outcomes of these r.v.s (as above),
we have
\begin{equation*}
    |h(\bld{D}) - h(\bld{E})| \,\, \leq \,\, U \cdot \TV(\bld{D}, \bld{E}).
\end{equation*}
\end{lemma}

\begin{lemma}\label{lem:TV and Hellinger}
    We have the following inequality between total variaion distance and Hellinger distance:
    $$
    \H^2(P,Q)\leq \TV(P,Q) \leq \sqrt{2}\H(P,Q).
    $$
\end{lemma}

The following chain rule for $\H^2$ distance was proved by \cite{Jayram2009} in obtaining a lower bound  for the communication complexity of the AND-operator.
See also the lecture notes by   \cite{wu2024chainrule} that  interprets  this result from a statistical point of view that we use here. This rule can be applied to general distributions (it does not require finite support).
\begin{theorem}[Jayram's chain rule]\label{lem:Jayram}
Consider two  joint distributions $P$ and $Q$ on $n$ random variables $\langle Y_1,\cdots Y_n\rangle$ factorized as:
    $$
    P = P_{Y_1} \cdot P_{Y_2|Y_{[2]}} \dots P_{Y_n|Y_{[n]}},
    $$
    and
        $$
    Q = Q_{Y_1} \cdot Q_{Y_2|Y_{[2]}} \dots Q_{Y_n|Y_{[n]}},
    $$
where $Y_{[j]}= (Y_1 \dots Y_{j-1})$ follows the joint distribution on the first $j-1$ random variables. Then,
$$
    \H^2 (P,Q) \quad \leq \quad 4 \cdot \sum_{j=1}^n \E_P[\H^2(P_{Y_j|Y_{[j]}},Q_{Y_j|Y_{[j]}})].
    $$
\end{theorem}
The key aspect of this result is that it holds for any {\em dependent} sequence of r.v.s. In the special case where the r.v.s $Y_1,\cdots Y_n$ are independent,
such a chain rule for $\H^2$ distance is folklore (see \cite{Gibbs2002Choosing}), and we actually have  $\H^2 (P,Q)  \leq  \sum_{j=1}^n \H^2(P_{Y_j},Q_{Y_j})$.
In our setting of stochastic problems, although the underlying item-distributions  are independent, the probing decisions made by a policy  depend on the history,  leading to a sequence of dependent r.v.s. Luckily, Jayram's rule give us a tool to bound the  Hellinger distance even for  dependent sequences.

\section{Weakly Monotone Problems with Discrete Distributions}
In this section, we prove our first main result (Theorem~\ref{thm:main strong-discrete}), which holds for any weakly monotone stochastic problem with a discrete distribution.

\subsection{Stability Lemma via  Hellinger Distance}
Here, we present our main technical result, which provides an improved ``stability'' bound on the expected objective of any policy $\sigma$ under two different distributions $\bD$ and $\bE$. The key point here is the use of Hellinger distance: the previous such result in \cite{agarwal2023semi} obtained an upper bound that depended linearly on  the TV-distance, which only yields $\tilde{O}(n\sqrt{T})$ regret.

\begin{lemma}\label{lem:chain rule for Hellinger}
    Given a stochastic problem $\cP$ and two discrete product distributions with the same support set $\bD=\{\calD_i\}_{i \in \calI}$ and $\bE=\{\calE_i\}_{i \in \calI}$, for any policy $\sigma$, we have
    $$
    f(\sigma | \bD) - f(\sigma | \bE) \leq 2\sqrt{2}  f_{\max} \cdot \sqrt{ \sum_{i \in \calI} {Q}_{i}(\sigma) \cdot\H^2(\calD_i,  \calE_i)  } ,
    $$
    where $Q_i(\sigma)$ is the probability that policy $\sigma$ probes item $i$.
\end{lemma}
The rest of this subsection proves this result.

\def\I{\ensuremath{{\cal I'}}\xspace}

\paragraph{Alternative View of Policy}
For the purpose of analysis only, we work with the following extended policy. We add $n$  dummy items, each of which corresponds to a zero-valued random variable. Let $\I$ denote the set of all $2n$ items.
We then extend policy $\sigma$ such that it probes exactly $n$ items in every root-leaf path. Specifically, after any leaf node of the original decision tree (with $z$ probed items) we add probes to $n-z$ dummy items. So, we can view policy $\sigma$ as performing exactly $n$ steps. For each step $j=1,\cdots n$, let $I_j\in \I$ denote the $j^{th}$ probed item and $V_j=X_{I_j}$ be its realization; we also use $Y_j=(I_j,V_j)$ to denote the joint outcome in step $j$. Note that both $I_j$ and $V_j$ are random variables. Any internal node in the decision tree corresponds to some  step $j\in [n]$ and is associated with  the state
$s_j=\langle i_1, v_1, \cdots i_{j-1}, v_{j-1}\rangle$
 consisting of all previously probed items $i_1,\cdots i_{j-1}$ with their realizations $v_1,\cdots v_{j-1}$. At each such internal node, the policy  specifies a probability distribution $\Pi_j(s_j)$ over  items (which  determines the next item to be probed).  All leaf nodes in the extended policy occur in step $n+1$ (after all $n$ probes). Each leaf node  is associated with a cost that is the same as the original leaf node in $T_\sigma$.
 Conditioned on any   state $\langle Y_1,\cdots Y_{j-1}\rangle$ in step $j$, the next r.v. $Y_j=(I_j,V_j)$ has the following distribution:
 \begin{equation}
     \label{eq:yj-distr} I_j \sim \Pi_j(\langle Y_1,\cdots Y_{j-1}\rangle) \quad \mbox{ and }\quad V_j\sim \calD_{I_j}.
 \end{equation}
We let $\bD^{\sigma}$  denote the resulting joint distribution of $Y_1,\cdots Y_n$. If the items follow distribution $\bE$, we use $\bE^{\sigma}$  for the distribution of $Y_1,\cdots Y_n$.
The key idea in our proof is to apply Theorem~\ref{lem:Jayram} on the  distributions $\bD^{\sigma}$ and $\bE^{\sigma}$ for the (adaptive) outcome sequence  $Y_1,\cdots Y_n$.
For any step $j$, we use the notation $Y_{[j]}=\langle Y_1,\cdots Y_{j-1}\rangle$ to denote all the outcomes in the first $j-1$ steps. For any state $s_j$ in step $j$, we  use $\Pi_j(s_j) = \{\pi_j (i|s_j): i\in \I\}$ to denote the probability distribution of the next item; so $\Pr[ I_j =i | Y_{[j]} = s_j]=\pi_j(i|s_j)$.

When  items follow distribution $\bD$, we have the following relationship for any item $i\in \I$:
\begin{equation}\label{eq:Q_i}
    Q_i(\sigma)= \Pr_{\bD}\left[\mbox{item $i$ probed by policy }\sigma \right] =\sum_{j=1}^n \Pr_{\bD^{\sigma}}[I_j =i] =\E_{\bD^{\sigma}}\left[ \sum_{j=1}^n\pi_j(i \mid Y_{[j]}) \right] .
\end{equation}
Using the fact that costs are associated with  leaf nodes that correspond to  states $Y_{[n+1]}=\langle Y_1,\cdots Y_n\rangle$,
\begin{equation}
    \label{eq:f-difference}
f(\sigma | \bD) - f(\sigma | \bE) = \E_{\bD^{\sigma}}[f_{\sigma, Y_{[n+1]}}] - \E_{\bE^{\sigma}}[f_{\sigma, Y_{[n+1]}}]
\leq f_{\max}\cdot \TV (\bD^{\sigma},\bE^{\sigma}) \leq \sqrt{2} f_{\max} \cdot \H(\bD^{\sigma},\bE^{\sigma}),\end{equation}
where the first inequality is by Lemma~\ref{lem:function-val-tv} and second inequality is by Lemma~\ref{lem:TV and Hellinger}.

It remains to bound the Hellinger distance $\H(\bD^{\sigma},\bE^{\sigma})$ between the two dependent distributions $\bD^{\sigma}$ and $\bE^{\sigma}$. To this end, we use the ``adaptive chain rule'' Theorem~\ref{lem:Jayram} and obtain:
\begin{equation}\label{eq:apply-chain-rule}
    \H^2(\bD^{\sigma},\bE^{\sigma}) \quad \leq\quad 4\cdot  \sum_{j=1}^n \E_{\bD^{\sigma}}\left[\H^2\left(\bD^{\sigma}_{Y_j|Y_{[j]}}, \bE^{\sigma}_{Y_j|Y_{[j]}}\right)\right].
\end{equation}
Fix any step $j\in [n]$ and state $Y_{[j]}=s$ in step $j$. Using the rule \eqref{eq:yj-distr} for the conditional distribution $Y_j|s$,  we have
\begin{equation}\label{eq:chain for mass}
\Pr_{\bD^{\sigma}}[Y_j = (i,v) \,|\, s]=\pi_j(i|s)\cdot \Pr_{\calD_i}[X_i=v] = \pi_j(i|s)\cdot \calD_i(v),\qquad \forall i\in \I \mbox{ and }v\in V,
\end{equation}
where $V$ denotes the support of all the r.v.s in $\I$. Similarly,
$$\Pr_{\bE^{\sigma}}[Y_j = (i,v) \,|\, s]=\pi_j(i|s)\cdot \calE_i(v),\qquad \forall i\in \I \mbox{ and }v\in V.$$
Hence,
\begin{align*}
\H^2\left(\bD^{\sigma}_{Y_j|s}, \bE^{\sigma}_{Y_j|s}\right) &= \frac{1}{2} \sum_{i \in \I} \sum_{v\in V } \left( \sqrt{\pi_j(i \mid s) \calD_i(v)} - \sqrt{\pi_j(i \mid s) \calE_i(v)} \right)^2\\
 &= \sum_{i \in \I} \pi_j(i \mid s) \cdot \left[ \frac{1}{2} \sum_{v\in V} \left( \sqrt{\calD_i(v)} - \sqrt{\calE_i(v)} \right)^2 \right] \,\,= \,\,\sum_{i \in \I} \pi_j(i \mid s) \H^2(\calD_i, \calE_i).
\end{align*}
$$\E_{\bD^{\sigma}}\left[\H^2\left(\bD^{\sigma}_{Y_j|Y_{[j]}}, \bE^{\sigma}_{Y_j|Y_{[j]}}\right)\right] = \E_{\bD^{\sigma}}\left[ \sum_{i \in \I} \pi_j(i \mid Y_{[j]}) \cdot \H^2(\calD_i, \calE_i)\right]=\sum_{i\in \I}   \H^2(\calD_i, \calE_i) \cdot \E_{\bD^{\sigma}}\left[    \pi_j(i \mid Y_{[j]}) \right].
$$
Adding over all steps and using \eqref{eq:apply-chain-rule},
\begin{align*}
 &\H^2(\bD^{\sigma},\bE^{\sigma})  \leq    4\cdot  \sum_{j=1}^n \E_{\bD^{\sigma}}\left[\H^2\left(\bD^{\sigma}_{Y_j|Y_{[j]}}, \bE^{\sigma}_{Y_j|Y_{[j]}}\right)\right]   = 4\cdot \sum_{j=1}^n \sum_{i\in \I}   \H^2(\calD_i, \calE_i) \cdot \E_{\bD^{\sigma}}\left[   \pi_j(i \mid Y_{[j]}) \right] \\
     & = 4\cdot  \sum_{i\in \I}   \H^2(\calD_i, \calE_i) \cdot \E_{\bD^{\sigma}}\left[ \sum_{j=1}^n  \pi_j(i \mid Y_{[j]}) \right] = 4\cdot  \sum_{i\in \I}   \H^2(\calD_i, \calE_i) \cdot Q_i(\sigma)= 4\cdot  \sum_{i\in \calI}   \H^2(\calD_i, \calE_i) \cdot Q_i(\sigma).
\end{align*}
The second to last equality used \eqref{eq:Q_i} and the last equality uses the fact that the Hellinger square distance between dummy r.v.s is zero by design. Combined with \eqref{eq:f-difference}, we obtain
$$f(\sigma | \bD) - f(\sigma | \bE) \le \sqrt{2} f_{\max} \cdot \H(\bD^{\sigma},\bE^{\sigma})\le 2\sqrt{2} f_{\max} \cdot\sqrt{\sum_{i\in \calI}   \H^2(\calD_i, \calE_i) \cdot Q_i(\sigma)},$$
which completes the proof.

\subsection{Bounding Regret using the Stability Lemma}\label{sec:thm discrete}
We now complete the proof of Theorem~\ref{thm:main strong-discrete}. We first state a sampling result that will be used in the online learning algorithm. This result follows fairly directly from techniques in \cite{GuoHT+21}; so  the proof is deferred to Appendix~\ref{app:sampling}.
\begin{restatable}{theorem}{thmsampling}
    \label{thm: Sample complexity}
    There is an algorithm $\ESD$ that, given $m \geq 5 \log(2T\delta^{-1}) $ i.i.d.\ samples from a distribution $\calD$ with finite support-size $k$, and parameter $\delta > 0$,  computes a distribution $\calE$ that satisfies the following properties with probability at least $1-\delta$:
\begin{itemize}
\item $\calE$ stochastically dominates $\calD$, and
\item the Hellinger square distance is $\H^2(\calE, \calD) = O\left(\frac{k\log^2(T/\delta)}{m}\right)$.
\end{itemize}
\end{restatable}

We  use the same meta algorithm as in \cite{agarwal2023semi}. The only change is in using the above sampling result (that bounds Hellinger distance) instead of their sampling result (that involved total-variation distance).
\begin{algorithm}
\caption{Online Semi-Bandit Algorithm}
\label{alg:greedy-ucb}
\begin{algorithmic}[1]
\For{$t = 1, \ldots T$}
\State for each item $i\in {\cal I}$, obtain distribution ${\calE}^t_i$ by running $\ESD$ with $\delta =\frac{1}{T^3}$ on all samples of r.v. $\varX_i$  observed so far.
\State obtain policy $\sigma^t$ by running the offline algorithm $\ALG$ on distribution $\bld{E}^t =\left\{{\calE}^t_i\right\}_{i=1}^n$.
\State run  policy $\sigma^t$ on realization $\bx^t$ and observe  the probed items.
\EndFor
\end{algorithmic}
\end{algorithm}

The analysis of this online algorithm follows the same high-level approach as \cite{agarwal2023semi}. However, some of the details differ because we utilize our stronger stability lemma.  We defer some proofs (Lemma~\ref{lem:stb:weak} and Lemma~\ref{lem:good-event}) to Appendix~\ref{app:missing}.

\begin{restatable}{lemma}{lemstbweak}\label{lem:stb:weak}
Consider a stochastic problem that is weakly monotone. Suppose that  $\bD=\{\calD_i\}_{i \in \calI}$ and $\bE=\{\calE_i\}_{i \in \calI}$ are product distributions
such that $\bD \sd \bE$.
If $\sigma$ is the policy returned by $\ALG(\bE)$
and $\sigma^*_\bD$ is an optimal  policy under $\bD$, then:
\begin{align*}
f(\sigma| \bD) - \alpha\cdot  f(\sigma^*_\bD| \bD) \, \, \leq \,\, 2\sqrt{2} f_{\max} \sqrt{ \sum_{i \in \calI} {Q}_{i}(\sigma) \cdot\H^2(\calD_i, \calE_i)  }
\end{align*}
where ${Q}_{i}(\sigma)$ denotes the probability that item $i$ is probed by policy $\sigma$ under distribution $\bld{D}$.
\end{restatable}
We will assume, without loss of generality, that we have observed at least $m \geq 5 \log(2T\delta^{-1}) $ samples from each random variable. This can be ensured by adding $5n \log(2T\delta^{-1}) $ special time periods and choosing for each time $i\in [n]$, a policy $\sigma^i$ that first probes item $i$; note that this  contributes at most $5n \log(2T\delta^{-1})  \cdot f_{max}$ to the total regret.

 We use $\bx^{t}$ to represent  the realization of all probed items at time $t$. For conciseness, we use $h^{t-1}$ to denote the {\it history} until time $t$; that is, $h^{t-1} = (\bx^1, \ldots, \bx^{t-1})$. Let $N_i^t$ be the number of times r.v. $X_i$ has been sampled before time $t$.  Note that $N_i^t$ is a random variable and depends on the history $h^{t-1}$. When needed, we will use $N_i^t(h^{t-1})$ to make this dependence explicit.

\begin{restatable}{lemma}{lemgoodevent}\label{lem:good-event}
With probability at least $1-\frac{n}{ T}$, for each $i \in[n]$ and $t\in [T]$,  we have that
\begin{itemize}
\item $ \calD_i\sd\calE_i^t$,
\item  $\H^2(\calE_i, \calD_i) = O\left(\frac{k\log^2(T)}{N^t_i}\right)$.
\end{itemize}
\end{restatable}

 Let   denote the {\it ``good'' event}   corresponding to the condition in Lemma~\ref{lem:good-event} holding for all item $i$ and period  $t$.

\paragraph{Bounding the regret as a sum over time $t$.} To bound the overall regret,  it suffices to bound the regret at each time $t\in [T]$, defined as:
\begin{align*}
R^t(h^{t-1}) \,\, &:= \,\, f(\sigma^t|\bD) - \alpha\cdot f(\sigma^*_{\bD}|\bD) ,
\end{align*}
where policy $\sigma^t=\ALG(\bE^t)$ and policy $\sigma^*_{\bD}$ is
the optimal policy for the stochastic problem instance (under the true distribution $\bD$). For each time $t\in [T]$ and history $h^{t-1}$, we apply  Lemma~\ref{lem:stb:weak} with distributions $\bD $ and  $ \bE^t(h^{t-1})$.  To keep notation simple, we use  function $g(m):=O(1) \cdot \frac{k\log^2(T)}{m} $ where the $O(1)$ term comes from Theorem~\ref{thm: Sample complexity}.
Note that  $\H^2(\calE_i, \calD_i) \le  g(N_i^t)$ and $\calD_i\sd \calE_i$  for each $i \in [n]$ because we assumed the good event $G$. Hence, Lemma~\ref{lem:stb:weak} implies that:
\begin{equation}
\label{eq:regret-t-stable}
R^t(h^{t-1})  \le  2\sqrt{2}  f_{\max} \sqrt{ \sum_{i \in \calI} {Q}_{i}(\sigma^t) \cdot\H^2(\calD_i, \calE_i^t)  } .
\end{equation}

Therefore, the total expected regret is:
\begin{align}
&\E\left[\sum_{t=1}^T R^t(h^{t-1})\right]  \,\, \le  \,\,   2\sqrt{2}  f_{\max} \sum_{t=1}^T \E_{h^{t-1}}\left[  \sqrt{\sum_{i \in \calI} {Q}_{i}(\sigma^t) \cdot g(N_i^t)}\right] \notag  \\
& \le   \,\,   2\sqrt{2} f_{\max} \sum_{t=1}^T\sqrt{ \E_{h^{t-1}}\left[  \sum_{i \in \calI} {Q}_{i}(\sigma^t) \cdot g(N_i^t)\right]}  \,\,  =  \,\, 2\sqrt{2}   f_{\max} \sum_{t=1}^T 1\cdot  \sqrt{ \E_{h^{t-1}}\left[  \sum_{i \in \calI} {Q}_{i}(\sigma^t) \cdot g(N_i^t)\right]} \notag \\
&\leq   \,\, 2\sqrt{2} f_{\max} \sqrt{T } \cdot \sqrt{  \sum_{t=1}^T \sum_{i \in \calI}\E_{h^{t-1}}\left[ {Q}_{i}(\sigma^t) \cdot g(N_i^t)\right]} \label{eq:exp-tot-R}
\end{align}
Above,  the second inequality is by Jensen's Inequality for the concave function $\sqrt{x}$, and the last inequality uses the
Cauchy-Schwartz inequality.

We now  show that for any item $i$,
\begin{align}
& \sum_{t=1}^T    \E_{h^{t-1}}\left[  {Q}_{i}(\sigma^t) \cdot  g(N_i^t)\right] = O\left(k \log^3 (T) \right). \label{eq:path-regret-sum}
\end{align}

\paragraph{Proving \eqref{eq:path-regret-sum} as a sum over decision paths.} We refer to the full history $h^T=\left(\bx^1,\cdots , \bx^T\right)$ of the algorithm as its {\em decision path}. The main idea in this proof is to  view the left-hand-side in \eqref{eq:path-regret-sum}  as a sum over decision paths rather than a sum over time. To this end, define $Z_i(h^T)$ as:
\begin{equation}
\sum_{t=1}^T  \1\left[\text{item $i$ probed by }\sigma^t(h^{t-1})\right]\cdot g(N^t_i(h^{t-1})).
\end{equation}

Above, $\1$ is the indicator function.
By linearity of expectation,  $\E_{h^T} \left[ Z_i(h^T) \right]$ equals
{\small \begin{align}
&\sum_{t=1}^T \E_{h^T} \left[ \1\left[i \text{  probed by  }\sigma^t(h^{t-1})\right]\cdot g(N^t_i(h^{t-1})\right] \notag \\
& =\,\,\sum_{t=1}^T \E_{h^{t-1}, \bx^t} \left[ \1\left[i \text{  probed by  }\sigma^t(h^{t-1})\right] \cdot g(N^t_i(h^{t-1}) \right] \notag \\
& =\sum_{t=1}^T \E_{h^{t-1}} \left[ g(N^t_i) \cdot \pr_{\bx^t} [i \text{  probed by  }\sigma^t]\right]  = \,\, \sum_{t=1}^T \E_{h^{t-1}} \left[  Q_{i}(\sigma^t)\cdot  g(N^t_i) \right],\label{eq:Zi}
\end{align}}
where the first equality uses the fact that event \{$i$   probed by  $\sigma^t$\}  only depends on $h^t = (h^{t-1}, \bx^t)$, the
second  equality uses the fact that $N^t_i$ only depends on $h^{t-1}$ and that $\bx^t$ is independent of $h^{t-1}$, and the last equality is by the definition of $Q_{i}(\sigma^t)$.

Therefore, it suffices to upper bound $\E_{h^T}[Z_i(h^T)]$, which we can prove in a per-realization manner. Indeed, $Z_i(h^T)$ equals:
\begin{align*}
  & \sum_{t=1}^T  \1\left[i\text{ probed by }\sigma^t(h^{t-1})\right] g(N^t_i)\,\, \le \,\,\sum_{t=1}^T g(t)  =O\left( k\log^2(T) \right) \sum_{t=1}^T \frac1{ t} = O\left(k\log^3(T)\right).
\end{align*}
The first inequality uses the fact that $N^t_i$ equals the number of probes of item $i$ in the first $t-1$ time steps: so $N^{t+1}_i=N^t_i+1$ whenever item $i$
is probed by $\sigma^t$. This completes the proof of  \eqref{eq:path-regret-sum}.

\paragraph{Final analysis.} Assuming the good event, using \eqref{eq:exp-tot-R} and \eqref{eq:path-regret-sum}, the total expected regret is at most
  $O\left( f_{\max} \left(\sqrt{n kT} \log^{\frac{3}{2}}(T) + n\log(T)\right ) \right)$. As the good event holds with probability at least $1-\frac{n}{T}$, its complement occurs with probability at most $\frac{n}{T}$, during which we incur at most $f_{\max} \cdot  T$ regret. So, the   final expected regret remains $O\left( f_{\max} \left(\sqrt{n kT} \log^{\frac{3}{2}}(T) + n\log(T)\right ) \right)$.

This completes the proof of Theorem~\ref{thm:main strong-discrete}.

\section{Strongly Monotone and General Distributions}\label{sec:general distribution} In this section, we prove our second result (Theorem~\ref{thm:main strong-general}), which handles strongly monotone problems with general distributions. Here, each random variable $X_i \sim \mathcal{D}_i$ is supported over the bounded interval $[0, 1]$ with no further structural assumptions placed on $\calD_i$. Moreover, we restrict attention to stochastic problems $\cP$ for which $\ALG$ is guaranteed to return an optimal policy: so we only obtain  $1$-regret bounds.

One approach to generalizing from discrete to general distributions is to consider $k$-threshold policies as done in~\cite{agarwal2023semi}. Such a policy is represented by a decision tree \( \sigma \), where each internal node is labeled by an item \( i \in [n] \) to probe, along with a sequence of thresholds \( \langle \tau_1, \dots, \tau_{k-1} \rangle \) that partition the support \([a, b]\) into $k$ intervals (each of which corresponds to a branch out of this node).
While this class of policies does capture some classical problems, it is fundamentally limited to deterministic, threshold-based policies.

To overcome this limitation and accommodate general, randomized policies, we use this threshold-based model as a conceptual stepping stone toward full generalization. In the following, we introduce a measure-theoretic framework termed the \emph{canonical stochastic problem}, drawing ideas from the canonical bandit framework (see Chapter 4.6 of \citep{lattimore2020bandit}).

Given  a product distribution $\bD = \{\calD_i\}_{i\in \calI}$, we augment it with $n$ dummy items, each corresponding to a zero-valued random variable. Let $\I$ denote the full set of $2n$ items. The sequential interaction between a policy and the environment produces an outcome sequence $Y_{[n+1]} = \langle I_1, V_1, \dots, I_n, V_n \rangle$. For each $j \in [n]$, we define the space $\Omega_j = ([2n] \times [0,1])^{j-1}$, a specific realization $s_j = \langle i_1, v_1, \dots, i_{j-1}, v_{j-1}\rangle \in \Omega_j$, and the associated Borel $\sigma$-algebra $\mathcal{F}_j = \mathcal{B}(\Omega_j)$. The random variables are formally defined via coordinate projections:
\[
I_j(s_{n+1}) = i_j \qquad \text{and} \qquad V_j(s_{n+1}) = v_j,
\]
where $s_{n+1} = \langle i_1, v_1, \dots, i_{n}, v_{n}\rangle$. We formally define our policies as follows.
\begin{definition}
A policy $\sigma$ is uniquely characterized by a sequence of decision rules $(\pi_j)_{j=1}^n$ and a cost function $f_{\sigma}(\cdot)$. Each $\pi_j$ is a probability kernel from $(\Omega_{j}, \mathcal{F}_{j})$ to $([2n], 2^{[2n]})$. To ensure that the policy probes any random variable at most once, we enforce $\pi_j(i \mid s_j) = 0$ for any $i \in s_j$.  The cost function $f_{\sigma} : \Omega_{n+1} \rightarrow [0, f_{\max}]$ depends strictly on the realized ordering and values of the non-dummy random variables. Finally, the problem instance $\cP$ may specify a set of constraints $\cC$ that restricts the class of admissible policies.
\end{definition}

As shown in Proposition 4.8 of \cite{lattimore2020bandit}, there is a "unique" joint probability measure on $(\Omega_{n+1},\mathcal{F}_{n+1})$, which finishes our construction. A more comprehensive discussion regarding the  validity of this  construction can be found there. Our goal is to provide a semi-bandit learning algorithm for such stochastic problems:
$$    \text{minimize} \quad f(\sigma|\bD) \quad  \text{subject to} \quad \sigma \in \cC.$$

\subsection{Stability Lemma under General Distributions}
A core challenge in this general setting is carefully handling the chain rule for the Hellinger distance under arbitrary measures. We adopt the same high level as in the discrete case Lemma~\ref{lem:chain rule for Hellinger}.  However we need to be careful how to handle \eqref{eq:chain for mass} in this case.
\begin{lemma}\label{lem: general chain rule}
    Given a stochastic problem $\cP$ and two product distributions $\bD=\{\calD_i\}_{i \in \calI}$ and  $\bE=\{\calE_i\}_{i \in \calI}$, for any policy $\sigma$, we have
    $$
    f(\sigma | \bD) - f(\sigma | \bE) \leq 2\sqrt{2}  f_{\max} \cdot \sqrt{ \sum_{i \in \calI} {Q}_{i}(\sigma) \cdot\H^2(\calD_i,  \calE_i)  }
    $$
\end{lemma}
\begin{proof}
Recall the notation from Lemma~\ref{lem:chain rule for Hellinger}. For each step
$j=1,\ldots,n$, let $I_j\in \I$ denote the $j$-th probed item, and let
$V_j=X_{I_j}$ be its realization. We write $Y_j=(I_j,V_j)$ for the joint
outcome at step $j$, and use $Y_{[j]}=\langle Y_1,\ldots,Y_{j-1}\rangle$
to denote the history of outcomes in the first $j-1$ steps. To compute the
Hellinger distance (see ~\cref{def:Hellinger}) for $Y_j\mid Y_{[j]}=s$, where $s$ is a in $\Omega_j$, we first specify a common dominating measure.

For each item $i$, let $\lambda_i=\calD_i+\calE_i$ be a measure on
$([0,1],\mathcal B([0,1]))$. With a slight abuse of notation, we use $\calD_i$
and $\calE_i$ to denote both the distributions and their induced probability
measures. By construction, both $\calD_i$ and $\calE_i$ are absolutely
continuous with respect to $\lambda_i$. Define $\lambda=\sum_{i\in\I}\lambda_i$. (Note that $\lambda$ is a not a probability measure. It is a common finite measure.) Then $\calD_i$ and $\calE_i$ are absolutely continuous with respect to
$\lambda$ for every $i\in\I$. Let $\rho$ denote the counting measure on $[2n]$, satisfying $\rho(B) = |B|$ for any $B \subseteq [2n]$. We define the common dominating measure $\mu=\rho\times\lambda$ on
$([2n]\times[0,1],2^{[2n]}\otimes\mathcal B([0,1]))$. This measure dominates
 $Y_j\mid s$ under both $\bD$ and $\bE$.

Let $d_i(v)=\frac{d\calD_i}{d\lambda}(v)$ and
$e_i(v)=\frac{d\calE_i}{d\lambda}(v)$ denote the Radon--Nikodym derivatives of
$\calD_i$ and $\calE_i$ with respect to $\lambda$. Let $p^{\bD}_{Y_j\mid s}$
be the Radon--Nikodym derivative of $Y_j\mid s$ under
environment $\bD$ with respect to $\mu$. Then, for every $i\in\I$ and
$v\in[0,1]$, we have
$$
p^{\bD}_{Y_j\mid s}(i,v)=\pi_j(i\mid s)d_i(v).
$$
Similarly, if $p^{\bE}_{Y_j\mid s}$ denotes the Radon--Nikodym derivative  of $Y_j\mid s$ under environment $\bE$ with respect to
$\mu$, then  for all $i\in\I$ and $v\in[0,1]$,
$$
p^{\bE}_{Y_j\mid s}(i,v)=\pi_j(i\mid s)e_i(v).
$$

Hence, expanding the definition of the Hellinger distance using these
densities, we have
\begin{align*}
\H^2\!\left(\bD^\sigma_{Y_j\mid s},\bE^\sigma_{Y_j\mid s}\right)
&=\frac12\int_{[2n]\times[0,1]}
\left(\sqrt{p^{\bD}_{Y_j\mid s}(i,v)}-\sqrt{p^{\bE}_{Y_j\mid s}(i,v)}\right)^2\,d\mu(i,v) \\
&=\frac12\sum_{i\in\I}\int_{[0,1]}
\left(\sqrt{\pi_j(i\mid s)d_i(v)}-\sqrt{\pi_j(i\mid s)e_i(v)}\right)^2\,d\lambda(v) \\
&=\sum_{i\in\I}\pi_j(i\mid s)\cdot\frac12\int_{[0,1]}
\left(\sqrt{d_i(v)}-\sqrt{e_i(v)}\right)^2\,d\lambda(v) \\
&=\sum_{i\in\I}\pi_j(i\mid s)\,\H^2(\calD_i,\calE_i),
\end{align*}
where the second equality follows from Fubini's theorem applied to the product
measure $\mu=\rho\times\lambda$.
    The remainder of the analysis proceeds identically to the proof of Lemma~\ref{lem:chain rule for Hellinger}.
\end{proof}

\subsection{Completing the Proof of Theorem~\ref{thm:main strong-general}}\label{app:main strong-general}

We provide a stability lemma for strong monotone problems. We also note that $\hat\calD$ is for analysis only. We will still use $\sigma$ returned by $\ALG$ on distribution $\bE$.
\begin{lemma}\label{lem:stb:strong}
Consider a stochastic problem that is strongly monotone. Suppose that  $\bD=\{\calD_i\}_{i \in \calI}$ $\bE=\{\calE_i\}_{i \in \calI}$ and $\hat\bD = \{\hat\calD_i\}_{i\in \calI}$ are product distributions
such that $\bD \sd \bE \sd \hat\bD$.
If $\sigma$ is the optimal policy returned by $\ALG(\bE)$
and $\sigma^*_\bD$ is an optimal  policy under $\bD$, then:
\begin{align*}
f(\sigma| \bD) -  f(\sigma^*_\bD| \bD) \, \, \leq \,\, 2\sqrt{2}  f_{\max} \sqrt{ \sum_{i \in \calI} {Q}_{i}(\sigma) \cdot\H^2(\calD_i, \hat \calD_i)  }
\end{align*}
where ${Q}_{i}(\sigma)$ denotes the probability that item $i$ is probed by policy $\sigma$ under distribution $\bld{D}$.
\end{lemma}
\begin{proof}
Similar to Lemma~\ref{lem:stb:weak}, we have:
\allowdisplaybreaks
\begin{align*}
& f(\sigma| \bD) -    f(\sigma^*_\bD| \bD) \notag \\ &= \,\, f(\sigma| \bD) - f(\sigma| \hat\bD) +f(\sigma| \hat\bD)- f(\sigma| \bE) +f(\sigma| \bE) -    f(\sigma^*_\bD| \bD) \notag \\
& \le \,\,  f(\sigma| \bD) - f(\sigma| \hat\bD)  + \OPT(\bE) -    f(\sigma^*_\bD| \bD) \\
&= \,\,  f(\sigma| \bD) - f(\sigma| \hat\bD)  +   \OPT(\bE) -    \OPT(\bD) \\ &\leq f(\sigma| \bD) - f(\sigma| \hat\bD) \leq 2\sqrt{2}  f_{\max} \sqrt{ \sum_{i \in \calI} {Q}_{i}(\sigma) \cdot\H^2(\calD_i,  \hat\calD_i)  }.
\end{align*}
where the first inequality is by strongly monotonicity and $\sigma$ is the optimal policy under $\bE$. The second inequality is by weakly monotonicity and the last inequality is by Lemma~\ref{lem: general chain rule}.

\end{proof}

\begin{lemma}\label{lem: shifted distribution sample}
Given any general distribution $\calD$ bounded over $[0,1]$, let $\calE$ by the distribution induced by Algorithm~\ref{alg:emp_stoc_dom}. There exists an auxiliary distribution $\hat\calD$ that satisfies the following properties with probability at least $1-\delta$:
\begin{itemize}
\item $\hat\calD$ stochastically dominates $\calE$, and
\item the Hellinger square distance satisfies $\H^2(\calD, \hat\calD) = O\left(\frac{\log^2(T/\delta)}{m}\right)$.
\end{itemize}
\end{lemma}
\begin{proof}

We also consider an auxiliary distribution $\bar\calD$ defined by the following CDF:
\[ \bar{F}(x) = \max\left\{0, F(x) - \sqrt{ \frac{2 F(x) (1-F(x))\log(2T/\delta)}{m-1} } - \frac{4\log(2T/\delta)}{m-1} \right\}. \]

By applying the same analysis as in Theorem~\ref{thm: Sample complexity}, we can verify that $\bar\calD$ is a valid probability distribution.
We now define the distribution $\hat{\mathcal{D}}$ via the CDF:
\[ \hat{F}(x) = \max\left\{0, \bar{F}(x) - \sqrt{ \frac{2F(x) (1-F(x))\log(2T/\delta)}{m-1} } - \frac{4\log(2T/\delta)}{m-1} \right\}. \]
By the same analysis, $\hat\calD$ is also a valid distribution. Moreover, a straightforward application of the stochastic dominance analysis from that same theorem establishes that $\calE \sd \hat\calD $ with probability $1-\delta$.

When bounding the Hellinger distance,  by invoking Lemma~\ref{lem:Hellinger Dom} with $c = \frac{\log(2T\delta^{-1})}{m-1}$, we obtain:
\[
\H^2(\calD, \bar\calD) = O\left(\frac{\log(T/\delta)}{m} \log\left(\frac{m}{\log(T/\delta)}\right)\right) \le O\left(\frac{\log^2(T/\delta)}{m}\right).
\]
Similarly, we have $\H^2(\bar\calD, \hat\calD) =O\left(\frac{\log^2(T/\delta)}{m}\right)$. Then by triangle inequality, we have $\H(\calD,\hat\calD) \leq \H(\calD,\bar\calD)+\H(\bar\calD,\hat\calD)= O\left(\sqrt{\frac{\log^2(T/\delta)}{m}}\right)$.
\end{proof}

Now, by deploying the same algorithmic framework (Algorithm~\ref{alg:emp_stoc_dom} and Algorithm~\ref{alg:greedy-ucb}). With the guarantees from Lemma~\ref{lem: shifted distribution sample} and Lemma~\ref{lem:stb:strong}, we can follow the same analytical framework as in Section~\ref{sec:thm discrete} to complete the proof of Theorem~\ref{thm:main strong-general}.

\begin{remark}
    In order to obtain the results for Remark~\ref{remark:alpha-monotone}, we only need to modify Lemma~\ref{lem:stb:strong} in order to account of the $\alpha$-approximate oracle (see Lemma~\ref{lem:alpha-stable}). The rest of the analysis follows  the same steps.
\end{remark}

\bibliographystyle{alpha}
\bibliography{ref}

\appendix
\newcommand{\savelemmaformat}{
  \let\oldthelemma\thelemma
  \renewcommand{\thelemma}{\arabic{section}.\arabic{lemma}}
}
\newcommand{\restorelemmaformat}{\let\thelemma\oldthelemma
}
\savelemmaformat

\section{Proof of Sampling Theorem~\ref{thm: Sample complexity}} \label{app:sampling}
\subsection{Technical Setup}
In this subsection, we present several useful results, which are needed for the proof of Theorem~\ref{thm: Sample complexity}.
\begin{lemma}[\cite{GuoHT+21} Lemma 6]\label{lem:Hellinger empirical}
    Let $\calD$ be a discrete distribution with support size $k$ and $\calG$ be its empirical distribution from $m$ samples. Then with probability $1-\delta$, we have:
    $$
    \H^2(\calD,\calG) = O\left(\frac{k\log 1/\delta}{m}\right).
    $$
    where $m$ is the number of samples to construct $\calG$.
\end{lemma}

The following lemma is a modification \footnote{We note that in the original paper, the shifted distribution is induced by a slightly different CDF. However, their analytical framework applies analogously to our formulation of the shifted distribution, yielding the same asymptotic bound.}  of \cite{GuoHT+21} Lemma 25 and 26.
\begin{lemma}\label{lem:Hellinger Dom}
     Let $\calD$ with CDF $F(x)$ be any distribution and $c \in (0,\frac{1}{4})$. Let $\hat{\calD}$ be its shifted distribution, which is defined by the following CDF $\hat F(x):=\max\{0, F(x) - \sqrt{ 2c F(x) (1-F(x)) } -4c \}$.  We have:
    $$
    \H^2(\calD,\hat{\calD}) = O\left(c \log\frac{1}{c}\right).
    $$
\end{lemma}

\proof{\it of \Cref{lem:Hellinger Dom}.}
Following their (\cite{GuoHT+21} Lemma 25) approach, we define $g(y) = y - \sqrt{2c y(1-y)} - 4c$. It is enough to show that there is a unique solution $G_l$ to $g(y) = 0$ over the interval $[0,1]$. And we have
$$
c \leq G_l \leq 8c.
$$
 The rest just follows the same analysis in their original lemma (part (a) (b) (c)).

First, observe that $g(y) < 0$ for any $y < 4c$.
$$
g(y) = 0 \Leftrightarrow y-4c=\sqrt{2cy(1-y)} \Leftrightarrow(1+2c)y^2-10cy+16c^2=0.
$$

It has only one valid solution in $[4c,1]$:
$$
G_l=c\frac{5+\sqrt{9-32c}}{1+2c} \leq 8c,$$
the inequality is due to the fact that \(c\in(0,1/4)\).

Thus,
\[
c\le G_l\le 8c.
\]
 \hfill \qedhere
 \endproof

\begin{lemma}[\cite{maurer2009empirical}]\label{thm:Bern}
    Let $X_1, X_2, \dots, X_m$ be i.i.d. random variables such that $\mathbf{E}[X_i] = \mu$, and $|X_i| \leq 1$. Then, we have with probability at least $1-\delta$:
\[
\left| \frac{1}{m}\sum_{j=1}^m X_j - \mu \right| \leq \sqrt{\frac{2V_m \log2/\delta }{m}} + \frac{7\log2/\delta}{3m}.
\]
where $V_m = \frac{1}{m-1}\sum_{j=1}^m (X_j -\frac{1}{m}\sum_{i=1}^m X_i)^2$ is the sample variance.
\end{lemma}

\begin{lemma}\label{lem:DKW-emp-Bernstein}
   Let $\calD$ (with CDF $F(\cdot)$) be any distribution over $\mathbb{R}$ and $\calG$ (with CDF $G(\cdot)$) be its empirical distribution from $m$ samples. With probability at least $1-\delta$ for any any $z$,
$$
\left|F(z)-G(z)\right|
\le \sqrt{
\frac{2G(z)(1-G(z))\log(2T\delta^{-1})}{m-1}}
+
\frac{4\log(2T\delta^{-1})}{(m-1)}
.
$$
where $m$ is the number of samples to construct $\calG$.
\end{lemma}

\proof{\it of \Cref{lem:DKW-emp-Bernstein}.}
    We use the same analysis from \cite{GuoHZ19} Lemma 5. We replace the standard Bernstein inequality with Lemma~\ref{thm:Bern} in below. This would finish the proof.
    It suffices to show that for all $z\in \mathbb{R}$ such that $F(z)$ is
a multiple of $1/m$, the following holds:
$$
\left|F(z)-G(z)\right|
\le \sqrt{
\frac{2G(z)(1-G(z))\log(2T\delta^{-1})}{m-1}}
+
\frac{9\log(2T\delta^{-1})}{3(m-1)}
.
$$
Then, the lemma follows because the bound on difference in the CDF of any value in the two
distributions suffers an extra additive factor of at most
$\frac{1}{m}
\le
\frac{\log(2T\delta^{-1})}{m}.$

Then fix any $z$ such that $F(z)$ is a multiple of $1/m$. We notice $\mathbb{E}[G(z)] = F(z)$, $|G(z)| \le 1$. Moreover, the sample variance of $G(z)$ is $\frac{m}{m-1}G(z)(1-G(z)).$ Applying the empirical Bernstein inequality (Lemma \ref{thm:Bern}), we have that with probability at least $1 - \frac{\delta}{m}$:
$$
\left| G(z) - F(z) \right| \le \sqrt{\frac{2G(z)(1-G(z)) \log(2m\delta^{-1})}{m-1}} + \frac{7\log(2m\delta^{-1})}{3(m-1)}.
$$
Finally, by taking a union bound over $m \in [T]$ and upper bounding $\log(2m\delta^{-1})$ by $\log(2T\delta^{-1})$, we finish the proof.
 \hfill \qedhere
\endproof

\subsection{Completing the proof of the Sampling Theorem}
\thmsampling*

The high-level idea of the proof is  from \cite{GuoHT+21}.
\begin{algorithm}
\caption{\textsc{EmpStocDom}}
\label{alg:emp_stoc_dom}
\begin{algorithmic}[1]
\State \textbf{Input:} Support set $[0,1]$, samples  $x_1, \ldots, x_m\sim \calD$, confidence parameter $\delta$
\State for $x \in \supp$, let $\hF(x) \leftarrow \frac1m\cdot  \sum_{j = 1}^m \1[x_j \leq x]$
\State let $c = \frac{\log(2T\delta^{-1})}{(m-1)}$ and $\epsilon(x) =  \sqrt{2c\hF(x)(1-\hF(x))} + 4c$.
\State set $E(x) = (\hF(x) - \epsilon(x))^+$ for all $x \in [0,1)$ and $E(1) =1 $.
\State \textbf{Output:} distribution $\calE$ over support $[0,1]$ with CDF $E(x)$.
\end{algorithmic}
\end{algorithm}

In this proof, we use $F(\cdot)$ to denote the cumulative distribution function (CDF) of the true distribution $\calD$. We use $G(\cdot)$ and $\calG$ for the empirical CDF and its corresponding distribution. Similarly, we use $E(\cdot)$ and $\calE$ for the shifted CDF and the distribution produced by Algorithm~\ref{alg:emp_stoc_dom}.

\textbf{Valid Distribution} We use a similar idea from \cite{liu2025improved} Lemma C.1.4. We first note that $E(x)\geq 0$ and $E(1)= 1$. It is suffice to show that $E(x)$ is non-decreasing. Since $G(\cdot)$ is non-decreasing, we only need to show $E(x)$ is non-decreasing as a function of $G(x)$.

We use the same notation in the proof of \Cref{lem:Hellinger Dom} and notice that here $c = \frac{4\log(2T\delta^{-1})}{m-1}$ in particular. For any $x$ such that $G(x) \le G_l$, we have $E(x) = 0$, and thus it is non-decreasing. Then for any $x$ such that $G(x) \ge G_l$, we let $y = G(x)$, and we have $E(x) = g(y) = y - \sqrt{2cy(1-y)} - 4c$.

In this case, we use $g'(y) = 1 - \frac{\sqrt{c}(1-2y)}{\sqrt{2y(1-y)}}$ to denote its derivative. When $y \ge \frac{1}{2}$, this is obviously non-negative, and thus $g(y)$ is non-decreasing. Then, when $y < \frac{1}{2}$, we show that $\sqrt{2y(1 - y)} > \sqrt{c}(1 - 2y)$. As $y \ge c$, it is enough to show $\sqrt{2y(1 - y)} > \sqrt{y}(1 - 2y)$. This is equivalent to showing $4y^2 - 2y - 1 < 0$ when $y < \frac{1}{2}$, which can be easily verified.

\textbf{$\calE$ stochastic dominates $\calD$:} By Lemma~\ref{lem:DKW-emp-Bernstein}, the following  event holds with probability at least $1-\delta$: for any $x \in [0,1]$,
\[
\left|F(x)-G(x)\right| \le \sqrt{\frac{2G(x)(1-G(x))\log(2T\delta^{-1})}{m-1}} + \frac{4\log(2T\delta^{-1})}{m-1}.
\]
Under this event, we have:
\[
F(x) \ge \max\left\{G(x) -\sqrt{\frac{2G(x)(1-G(x))\log(2T\delta^{-1})}{m-1}} - \frac{4\log(2T\delta^{-1})}{m-1},\, 0\right\}= E(x), \quad \forall x\in [0,1].
\]
This guarantees that $\calD \sd \calE$.

\textbf{Bound Hellinger Distance} We now bound the Hellinger distance between $\calG$ and $\calE$. By invoking Lemma~\ref{lem:Hellinger Dom} with $c = \frac{\log(2T\delta^{-1})}{m-1}$, we obtain:
\[
\H^2(\calG, \calE) = O\left(\frac{\log(T/\delta)}{m} \log\left(\frac{m}{\log(T/\delta)}\right)\right) \le O\left(\frac{\log^2(T/\delta)}{m}\right).
\]
Separately, Lemma~\ref{lem:Hellinger empirical} guarantees that the empirical bound $\H^2(\calD,\calG) = O\left(\frac{k\log(1/\delta)}{m}\right)$ holds with probability at least $1-\delta$. Finally, applying a union bound over these high-probability events, the triangle inequality for the Hellinger distance yields:
\[
\H(\calE, \calD) \le \H(\calE,\calG) + \H(\calG, \calD) = O\left(\sqrt{\frac{k\log(1/\delta)}{m}} + \sqrt{\frac{\log^2(T/\delta)}{m}}\right) = O\left(\sqrt{\frac{k\log^2(T/\delta)}{m}}\right),
\]
which completes the proof.
\hfill \qedhere
\endproof

\section{Other Missing Proofs}\label{app:missing}

\lemstbweak*
\proof
We have:
\begin{align*}
& f(\sigma| \bD) - \alpha\cdot  f(\sigma^*_\bD| \bD) \notag \\ &= \,\, f(\sigma| \bD) - f(\sigma| \bE) +f(\sigma| \bE) - \alpha\cdot  f(\sigma^*_\bD| \bD) \notag \\
& \le \,\,  f(\sigma| \bD) - f(\sigma| \bE)  + \alpha\cdot \OPT(\bE) - \alpha\cdot  f(\sigma^*_\bD| \bD) \\
&= \,\,  f(\sigma| \bD) - f(\sigma| \bE)  + \alpha\cdot \OPT(\bE) - \alpha\cdot  \OPT(\bD) \\ &\leq f(\sigma| \bD) - f(\sigma|  \bE) \leq2\sqrt{2}  f_{\max} \sqrt{ \sum_{i \in \calI} {Q}_{i}(\sigma) \cdot\H^2(\calD_i,  \calE_i)  } .
\end{align*}
where the first inequality uses the $\alpha$-approximation ratio of the offline algorithm, the second inequality is by weak  monotonicity and the last inequality is by Lemma~\ref{lem:chain rule for Hellinger}.
\hfill \qedhere
\endproof

\lemgoodevent*
\proof
For any $j \in [n]$ and $t\in [T]$, let $B_j^t$ denote the event that the stated condition {\em does not} hold for $j$ and $t$. Using Theorem~\ref{thm: Sample complexity} with $\delta = 1/T^3$, we obtain for any $j\in [n]$ and $t\in [T]$,
\[
\pr\left(B_j^t \ \wedge \ N_j^t = m\right) \le  \delta,\qquad \forall m\ge 1.
\]
So, for any $j$ and $t$,  we have by union bound,
\[\pr(B_j^t)  \leq \sum_{m=1}^T \pr\left(B_j^t \ \wedge \ N_j^t = m\right) \leq \sum_{m=1}^T \delta = \delta T.\]
Again, by union bound,
\[\pr\left(\vee_{j=1}^n \vee_{t=1}^T  B_j^t\right)  \leq \sum_{j=1}^n \sum_{t=1}^T  \pr(B_j^t)  \le nT^2\delta\le \frac{n}{T}.\]
This completes the proof.
\hfill \qedhere
\endproof

\begin{lemma}\label{lem:alpha-stable}
Consider a stochastic problem that is strongly monotone. Suppose that  $\bD=\{\calD_i\}_{i \in \calI}$ $\bE=\{\calE_i\}_{i \in \calI}$ and $\hat\bD = \{\hat\calD_i\}_{i\in \calI}$ are product distributions
such that $\bD \sd \bE \sd \hat\bD$.
If $\sigma^*_\bD$ is an optimal  policy under $\bD$ and $\sigma$ is the $\alpha-$approximate policy returned by $\ALG(\bE)$, which is policy-wise monotone~\eqref{eq:policy-wise-monotone}, then:
\begin{align*}
f(\sigma| \bD) -  \alpha f(\sigma^*_\bD| \bD) \, \, \leq \,\, 2\sqrt{2}  f_{\max} \sqrt{ \sum_{i \in \calI} {Q}_{i}(\sigma) \cdot\H^2(\calD_i, \hat \calD_i)  }
\end{align*}
where ${Q}_{i}(\sigma)$ denotes the probability that item $i$ is probed by policy $\sigma$ under distribution $\bld{D}$.
\end{lemma}
\begin{proof}
Similar to Lemma~\ref{lem:stb:weak}, we have:
\allowdisplaybreaks
\begin{align*}
& f(\sigma| \bD) -   \alpha  f(\sigma^*_\bD| \bD) \notag \\ &= \,\, f(\sigma| \bD) - \alpha f(\sigma| \hat\bD) + \alpha f(\sigma| \hat\bD)- \alpha f(\sigma| \bE) +
\alpha f(\sigma| \bE) -    \alpha f(\sigma^*_\bD| \bD) \notag \\
& \le \,\,  f(\sigma| \bD) - \alpha f(\sigma| \hat\bD)  + \alpha \OPT(\bE) -    \alpha f(\sigma^*_\bD| \bD) \\
&= \,\,  f(\sigma| \bD) - \alpha f(\sigma| \hat\bD)  +  \alpha \OPT(\bE) - \alpha     \OPT(\bD) \\ &\leq f(\sigma| \bD) - \alpha f(\sigma| \hat\bD) \leq 2\sqrt{2}  f_{\max} \sqrt{ \sum_{i \in \calI} {Q}_{i}(\sigma) \cdot\H^2(\calD_i,  \hat\calD_i)  }.
\end{align*}
where the first inequality is by the assumptions in Remark~\ref{remark:alpha-monotone}. The second inequality is by weakly monotonicity and the last inequality is by Lemma~\ref{lem: general chain rule}.

\end{proof}

\section{Standard Results}\label{app:standard results}

In this section, we present some standard measure theory results.

\begin{definition}[\cite{lattimore2020bandit}]

    Let $(X, \mathcal{A})$ and $(Y, \mathcal{B})$ be measurable spaces. A probability kernel from $X$ to $Y$ is a mapping $\kappa: X \times \mathcal{B} \to [0, 1]$ satisfying the following two conditions:

\begin{enumerate}
    \item For every fixed set $B \in \mathcal{B}$, the mapping
$$
    x \mapsto \kappa(x, B)
$$
    is an $\mathcal{A}$-measurable function from $X$ into $[0, 1]$.

    \item For every fixed point $x \in X$, the mapping
$$
    B \mapsto \kappa(x, B)
$$
    is a probability measure on $(Y, \mathcal{B})$.
\end{enumerate}
\end{definition}

\begin{definition}[\cite{royden2010real}]
Let $(X, \mathcal{M})$ be a measurable space, and let $\mu$ and $\nu$ be two measures on this space. We say that $\nu$ is absolutely continuous with respect to $\mu$ (denoted as $\nu \ll \mu$) if for every measurable set $E \in \mathcal{M}$:
$$
\mu(E) = 0 \implies \nu(E) = 0
$$

\end{definition}

\begin{theorem}[Radon-Nikodym Theorem \cite{royden2010real}]\label{thm:RN}
     Let $(X, \mathcal{M}, \mu)$ be a $\sigma$-finite measure space and $\nu$ a $\sigma$-finite measure defined on the measurable space $(X, \mathcal{M})$ that is absolutely continuous with respect to $\mu$. Then there is a nonnegative function $f$ on $X$ that is measurable with respect to $\mathcal{M}$ for which
$$
\nu(E) = \int_E f \, d\mu \text{ for all } E \in \mathcal{M}.
$$
$f$ is called the Radon-Nikodym  derivative of $\nu$ with respect to $\mu$. The function $f$ is unique in the sense that if $g$ is any nonnegative measurable function on $X$ that also has this property, then $g = f$ almost everywhere with respect to $\mu$.
\end{theorem}
\end{document}